\documentclass{article}

\usepackage[preprint]{neurips_2023}

\usepackage[utf8]{inputenc} % allow utf-8 input
\usepackage[T1]{fontenc}    % use 8-bit T1 fonts
\usepackage{hyperref}       % hyperlinks
\usepackage{url}            % simple URL typesetting
\usepackage{booktabs}       % professional-quality tables
\usepackage{amsfonts}       % blackboard math symbols
\usepackage{nicefrac}       % compact symbols for 1/2, etc.
\usepackage{microtype}      % microtypography
\usepackage{xcolor}         % colors
\usepackage{graphicx}
\usepackage{float}
\usepackage{booktabs} % For better table formatting
\usepackage{multirow} % Required for multi-row cells
\usepackage{amsmath} % for \DeclareMathOperator

\title{Evaluating Brain-Inspired Modular Training in Automated Circuit Discovery for Mechanistic Interpretability}

\author{%
  Jatin Nainani  \\
  University of Massachusetts Amherst\\
  \texttt{jnainani@umass.edu} \\
}

\begin{document}

\maketitle

\begin{abstract}

  Large Language Models (LLMs) have experienced a rapid rise in AI, changing a wide range of applications with their advanced capabilities. As these models become increasingly integral to decision-making, the need for thorough interpretability has never been more critical. Mechanistic Interpretability offers a pathway to this understanding by identifying and analyzing specific sub-networks or 'circuits' within these complex systems. A crucial aspect of this approach is Automated Circuit Discovery, which facilitates the study of large models like GPT4 or LLAMA in a feasible manner. In this context, our research evaluates a recent method, Brain-Inspired Modular Training (BIMT), designed to enhance the interpretability of neural networks. We demonstrate how BIMT significantly improves the efficiency and quality of Automated Circuit Discovery, overcoming the limitations of manual methods. Our comparative analysis further reveals that BIMT outperforms existing models in terms of circuit quality, discovery time, and sparsity. Additionally, we provide a comprehensive computational analysis of BIMT, including aspects such as training duration, memory allocation requirements, and inference speed. This study advances the larger objective of creating trustworthy and transparent AI systems in addition to demonstrating how well BIMT works to make neural networks easier to understand.
\end{abstract}

\section{Introduction}

The primary idea of Mechanistic Interpretability is that we can find sparse, sub-networks (known as circuits) in a large network, that can be analyzed to understand their functionality. This decomposition of networks into circuits corresponding to their functionalities helps in interpretability. 

First documented by Olah et al. [5] for image circuits, this idea of zooming into the network to understand its inner workings is drawn from cellular biology. This paper first draws direct analogies from the claims about cells by Schwann and their speculative claims about neural networks. Each claim argues in favor of understanding the black box model by finding circuits. The claims are supported by examples from InceptionV1, where they find neurons and sub-graphs that respond to certain inputs. They demonstrate how meaningful features exist in neural networks by showcasing the presence of curve detectors and high-low frequency detectors. They also show how early curve detectors combine to form more complex shape detectors forming the circuits in question. Their last claim suggests that the features and circuits analyzed are similar across different models and tasks. This claim, if true, will help in generalizing our methodology for interpreting neural networks. 

Interpretability, which is central to this research, is about understanding the internal processes of a model. It's different from explainability, which is more about making the model's outputs understandable. Modular networks are preferred in interpretability because they allow for easier examination of specific parts of the network, offering insights into how the model makes decisions.

A major challenge in this field, as noted by Conmy et al. [2], is the difficulty in applying interpretability techniques to large, state-of-the-art models like GPT4 or LLAMA. This is why automating the process of finding these circuits is important - it makes studying large models more feasible. By analyzing circuits within a neural network, we can learn which features are important for certain tasks, find potential biases, and understand better how the model functions. This research aims to make such analysis easier and more accessible, contributing to making neural networks more transparent and reliable.

Biological neural networks (e.g., brains) differ from artificial ones in that biological ones restrain neuronal connections from being local in space, leading to anatomical modularity. Motivated by this observation, Liu et al. [1] proposed brain-inspired modular training (BIMT) to facilitate the modularity and interpretability of artificial neural networks. The idea is to embed neurons into a geometric space and minimize the total connection cost by adding the connection cost as a penalty to the loss function and swapping neurons if necessary. Several math and machine learning datasets show that BIMT can put functionally relevant neurons close to each other in space, just like brains. One of the claims of this research suggests that BIMT will aid other Mechanistic Interpretability techniques. 

In this paper, we rigorously evaluate Brain Inspired Modular Training (BIMT) positioned to enhance mechanistic interpretability. Our research primarily contributes to the field in the following ways:

\begin{itemize}
    \item This research is the first to undertake a comparative analysis of BIMT specifically for Automated Circuit Discovery, providing novel insights into its effectiveness.
    \item We demonstrate that BIMT significantly improves the efficiency and quality of Automatic Circuit Discovery. This advancement is pivotal as manual mechanistic interpretability methods are often too slow and impractical for large-scale models.
    \item Through extensive testing, we establish that BIMT outperforms existing models in several key areas: quality of discovered circuits, time efficiency in discovery, and the sparsity of the circuits. This comparative analysis offers a comprehensive view of BIMT's superiority in circuit discovery.
    \item We document detailed computational metrics for BIMT. This includes an analysis of the training duration in comparison with other models, memory allocation requirements, and inference speed. Such metrics are crucial for understanding the practical implications and scalability of BIMT.
\end{itemize}

\begin{figure}
    \centering
    \includegraphics[scale=0.5]{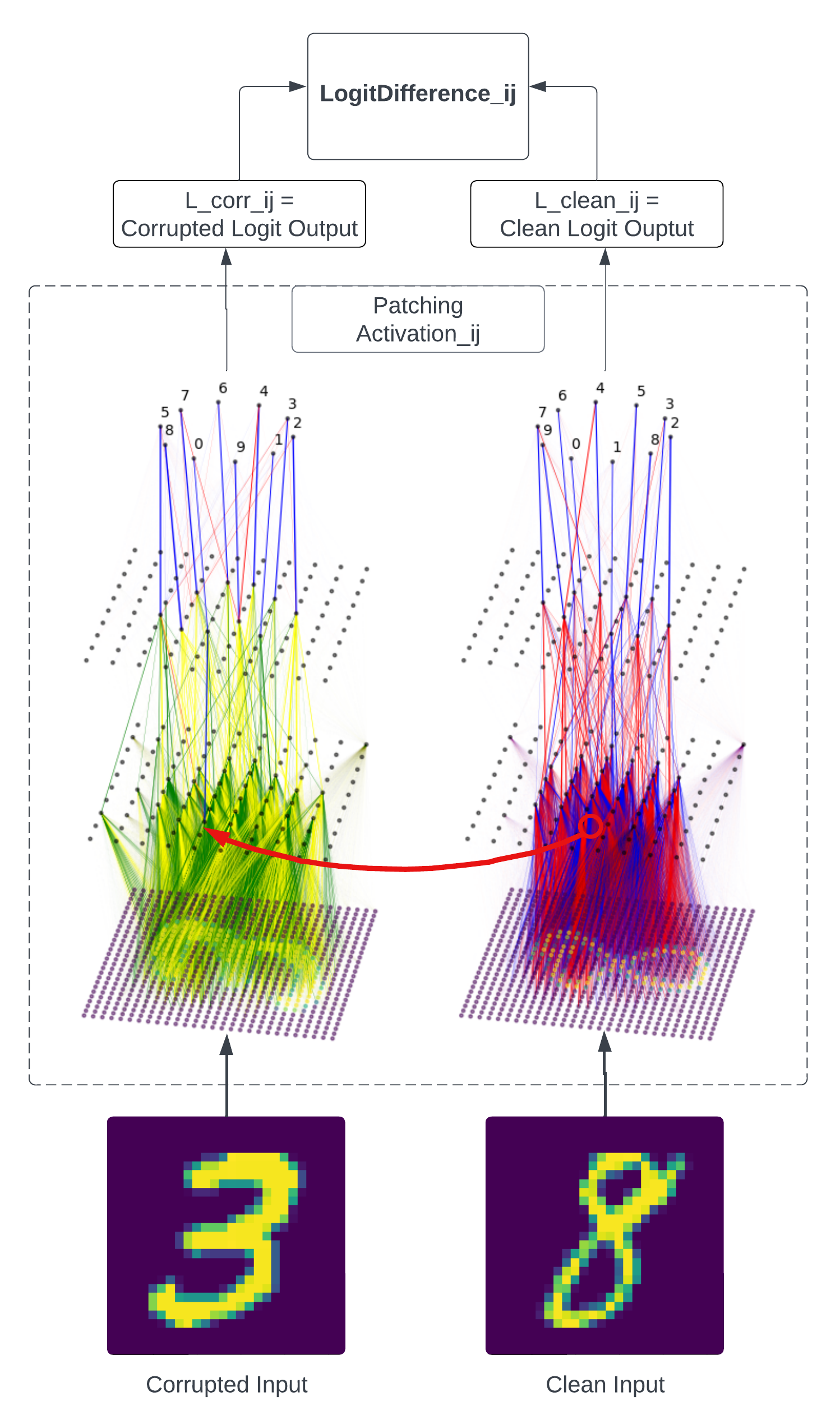}
    \caption{Demonstration of Activation Patching on Modular Trained Network. A clean input is given to the model to produce the expected behavior (detection of digit 8). The model is then given a corrupted input (digit 3) which leads to corrupted output. Weight connections that are blue and red correspond to the clean run, whereas green and yellow correspond to the corrupted run. We then iteratively copy activations from the clean network to the corrupted network. If replacing an activation $i$ on layer $j$ reduces the difference between $L\_clean_{ij}$ and $L\_corr_{ij}$, it means that neuron[i,j] is a relevant part of the circuit for this task. The figure shows the patching of neuron 47 at layer 1 to the corrupted run - represented by the blue and red lines at that neuron. This patched layer propagates as normal and produces a different logit output.}
    \label{fig:logits}
\end{figure}

\section{Research Design}

This section covers the theoretical aspects and motivations of the research. It first defines what Brain Inspired Modular Training is, and the training regimes we are investigating. It sets up two research questions that define the flow of the paper. It then focuses on setting up the fundamentals in methodologies and metrics that allow us to design experiments to answer the questions we defined.

\subsection{BIMT}

Liu et al. [1] propose a training regime for neural networks to make them more modular and interpretable. This idea, directly inspired by brains, aims to embed neurons in a geometric space and adds a distance-dependent loss function to penalize the length of each neuron connection. They suggest that training neural networks to form modules that we can directly see, will aid other mechanistic interpretability techniques. 

The proposed methodology for Brain-Inspired Modular Training is a mixture of three ideas: L1 + Local + Swap. L1 simply represents the L1 regularization.

\subsubsection{Neural Networks}

Finally, with the methods suggested by Liu et al. [1], we have 5 different kinds of training regimes.

\begin{table}[h!]
\centering
\caption{Summary of model training regimes under comparison}
\label{tab:models}
\begin{tabular}{@{}ll@{}}
\toprule
Model               & Description                                             \\ \midrule
Vanilla             & Dense Network with No Regularization \\
L1                  & Only L1 Regularization model                            \\
L1 + Local          & L1 Regularization that encourages locality              \\
L1 + Swap           & L1 regularization with periodic swaps                   \\
L1 + Local + Swap   & \textbf{BIMT}   \\ \bottomrule
\end{tabular}
\end{table}

These five will be our primary system for experimentation and comparison.

\subsection{Research questions}

In this research, we investigate the effect of modularity on automated mechanistic interpretability. Specifically, we look for the five training regimes that change the results of Automated Circuit Discovery. The idea of automating the discovery of circuits was proposed by Conmy et al. [2]. 

The biggest critique of mechanistic interpretability as a field is that manual circuit discovery and experimentation require efforts and time that are not feasible for interpreting large language models like GPT4, llama2, etc. Most papers on mechanistic interpretability focus on GPT2-small for their experimentation. This is the reason why we focus on evaluating BIMT for automated techniques. 

The idea is not to explain the trained model itself but to amend techniques that can help researchers to test their hypotheses on discovered circuits faster. 

\subsubsection{Research Question 1: How do networks trained to be modular affect automatic circuit discovery}

% This is our primary question, evaluation of BIMT for automated circuit discovery. As the paper by Liu et al. [1] mainly explored toy models, we hypothesize that modularity doesn't interact with automated circuit discovery in any way, and is meant for more manual techniques. If modular trained networks can produce circuits with better accuracy and lower sparsity than other models in question, the hypothesis will be falsified.

\subsubsection*{Hypothesis 1: Modularity only helps visualize small networks and contributes marginally to circuit discovery}

This is our primary question, evaluation of BIMT for automated circuit discovery. As the paper by Liu et al. [1] mainly explored toy models, we hypothesize that modularity doesn't interact with automated circuit discovery in any way, and is meant for more manual techniques. If modular trained networks can produce circuits with better accuracy and lower sparsity than other models in question, the hypothesis will be falsified.

\subsubsection*{Hypothesis 2: Modularity, under optimal conditions, narrows down search space for automated circuit discovery, speeding up the process}

The point of automated circuit discovery is to speed up the process of reverse engineering in large language models. Even if the circuits discovered between the models in comparison are similar, if the modular network achieves the circuit much faster than others, it will be helpful. The reason for this hypothesis comes from the assumption that the circuits that automated discovery are trying to find might already be segregated by modular training. If the circuit discovery times between models are relatively similar, the hypothesis is falsified. If modular networks are faster at finding the circuits, we can change the number of swaps (to reduce modularity) to confirm the hypothesis. 

\subsubsection*{Hypothesis 3: Modular trained networks hamper the ability to find optimal circuits}

The modules formed by modular training are done by projecting neurons on a geometric space and penalizing the neurons based on the distance from each other in a layer. On the other hand, circuit discovery looks for neurons that show significant contributions to a specific sub-task (detection of circles for digit classification for MNIST). These may be two inherently different ways of approaching interpretability. Conversely, it can also be the case that, modular training is grouping the neurons that contribute towards a specific sub-task, as it is swapping neurons that fire together to be closer. If that is not the case, then modular networks can potentially harm the ability of circuit discovery as it distorts the network structure. If we notice a decrease in metrics and/or an increase in training time for modular networks, then the hypothesis can be considered further. In other cases, it is falsified.

\subsubsection{Research Question 2: How does the computational efficiency vary with respect to modularity?}

% The modular training process needs to store the coordinates of each neuron in a layer to calculate distance-based metrics. This might lead to much higher memory requirements for the training process. If the memory requirements for training for a model with local regularization and L1-only regularization are similar, then the hypothesis is falsified.

\subsubsection*{Hypothesis 1: Modular training has higher memory requirements because of the multiple coordinate values needed to be stored}

The modular training process needs to store the coordinates of each neuron in a layer to calculate distance-based metrics. This might lead to much higher memory requirements for the training process. If the memory requirements for training for a model with local regularization and L1-only regularization are similar, then the hypothesis is falsified.

\subsubsection*{Hypothesis 2: Modular training takes much longer than other models because of the additional distance calculation}

The authors of the principal paper by Liu et al [1] state that swaps take $O(nkL)$ computations. While a swap is done only after approximately 200 iterations each, even the distance-based regularization needs additional computation compared to L1 regularization. This might lead to a higher training time for modular networks. We can easily falsify the hypothesis if there is no significant difference in training times.

\subsubsection*{Hypothesis 3: Inference times for models in questions have no significant difference, as inference only needs a forward pass}

Mechanistic Interpretability aims to reverse engineer networks that are "already trained". So the inference times matter much more than training times. As the forward pass of a modular network doesn't need to worry about swapping or distance calculation, I think there will be no difference in inference times for models of comparative sizes. If we notice a rise in inference times over multiple inputs, the hypothesis is falsified.

\subsection{General procedure for Mechanistic Interpretability research}

The work by Conmy et al. [2] identifies a workflow for finding and interpreting circuits that are followed by most researchers in the field. The paper breaks down the process of finding sub-networks in the following steps:
\begin{enumerate}
    \item Look for a specific task or behavior of the model, choose a data set where the task is displayed, and a metric for the model's performance on the task. 
    \item Decide the depth or granularity of interpretation needed, creating a computational graph.
    \item Iteratively run patching experiments and remove unnecessary connections from the graph.
    \item Once a circuit is isolated, a researcher can test their hypothesis for its functionality. 
\end{enumerate}

The paper by Conmy et al. [2] provides a way to automate step 3 in this process. This is done by modifications of an interpretation technique known as - recursive activation patching. For this particular type of MI technique, we will be heavily relying on the concept of activation patching which finds its traces from research by Meng et al. [3].

Our focus on the evaluation and empirical analysis is the third step in the process. We aim to see how modular training helps the patching experiments. The motivation behind this is the computational complexity of iterating through the graph is quite high [2].

\subsection{Quantifying Interpretability}

Quantifying interpretability is difficult, as it lacks a system that could be held to a standard of falsifiability. The idea of circuits makes evaluating interpretation feasible. If we can understand the workings of a sub-network, we should be able to reliably predict the outputs of it under different changes. This reliability forms the quantifying factor in MI.  

We have borrowed two general, high-level metrics for the evaluation of circuits from Automated Circuit Discovery by Conmy et al. [2]. 

\begin{itemize}
    \item Q1 - Is the circuit representing the underlying algorithm implemented by the neural network?
    \item Q2 - Does the circuit contain neurons that are not relevant to the task/behavior in question?
\end{itemize}

For Q1, we will rely on the loss of the task. If the subgraph can achieve relative performance on the specific task compared to the original model, then we can say that it implements the underlying algorithm. For comparison tasks, we will evaluate the difference between the logits of the model (Section ). 

For Q2, we will calculate the number of edges in the circuit. A Sparser circuit that can achieve the same or better metric for Q1 is always preferred. The idea behind this metric is that a smaller graph is easier to interpret. 

\subsection{Recursive Activation Patching}

This is the primary mechanistic interpretability method we are investigating. The method for activation patching, with Gaussian noise corruption, was first introduced by Meng et al. [3]. It serves as a technique to identify the neuron activations that contribute most significantly to determining the outputs between two inputs that differ in a key manner [5]. 

We start with a clean input (which produces the expected behavior) and a corrupted input (which does not produce the expected behavior) which produces a clean and corrupted output respectively. These two should ideally be similar only differing in a key detail that links to the behavior under question. We then patch activations from the clean run to the corrupted run to find the activations that can change the corrupted output to the clean output. This type of activation patching is Causal tracing [5]. We can also perform resample ablation by patching activations from the corrupted run to the clean run, but this method is prone to missing significant neurons when the model has redundancy [5]. 

The recursive part of the method was inspired by Conmy et al. [2], which lets it iteratively assess the significance of each activation. For each activation, we patch the clean activation to the corrupted one. Then perform a forward pass through the model to evaluate the metric for significance. 

\subsubsection{Logit Difference}
Logit Difference is a popular method to evaluate MI techniques. We can understand the concept with an example:

For one task/behavior, we assume the model to be optimal. For the example demonstrated in Figure 1, we say that a clean input MNIST model can optimally look for circles (the behavior in question) and classify digits like 6,8,9,0 accurately. In this case, our clean input will be a digit like 8, and our corrupted input will be a digit that is structurally similar but differs concerning the behavior like digit 3. 

It will have a logit output for the clean run, say, $L1$. This is our clean run. Then we provide the model with a corrupted input and we iteratively patch activations from the clean run and store their logit outputs $L2_{ij}$ for the $i^{th}$ activation in layer $j$. If an activation I from the clean run can reduce the difference $L1 - L2_{Ij}$, we can say that it is important in producing the optimal output. And thus, we are looking for activation patches with lower Logit Difference. We take an L2 norm of the logit difference output to create a robust variable representing the gap between the corrupted and clean outputs as shown in equation 6.

\begin{align}
\text{LogitDifference}_{ij} = \left\lVert \text{L\_corr}_{ij} - \text{L\_clean}_{ij} \right\rVert_2
\end{align}

Equation 7 defines the optimization problem of selecting the top K activations from each layer that minimize the logit difference. 

\begin{align}
\text{top\_K\_neurons}_{j} = \underset{i \in \text{neurons of layer} j}{\mathrm{arg\,min}^K} \left( \text{LogitDifference}_{ij}  \right)
\end{align}

\subsection{Other computational metrics}

To evaluate other parts of the research questions, we define additional metrics. 

\begin{itemize}
    \item Speed of circuit discovery - To measure the time efficiency of recursive activation patching in circuit formation. This ties back to the hypothesis that module formation reduces search space in MI techniques, potentially speeding up the process.
    \item Disk Space - the memory of each model file after training
    \item Training time - time taken to complete training 
    \item Inference time - average time of inference for one data point
    \item Memory Allocation - use of "torch.cuda.memory\_allocated" and "torch.autograd.profiler".
\end{itemize}

\section{Empirical Procedure}

This section sets up the implementation details of the experiments used to gather data. We narrowed down our focus to a specific type of neural network and task to create a fair comparison. Two behaviors (or tasks) are then selected for circuit discovery. The steps for recursive activation patching are formalized concerning the implementation.

\subsection{Data and Task for Original Network}

Before we start with the process of discovering a circuit, we need to define the task of the network we are trying to interpret. Hoping to stay faithful to the architectural decisions of BIMT and for fast feedback loops, we perform our experiments on the MultiLayer Perceptron for the task of digit classification on the MNIST dataset. 

\subsection{Task for Circuit}

A digit classification model considered is trained on images of handwritten digits. Specifically, as we will be using MNIST as our dataset, it is made of digits - 0,1,2,3,4,5,6,7,8,9. The objective of the experiment is to find a circuit in the overall model that detects circles in the input. A hypothetical sub-network of the original model that detects complete circles will demonstrate significant activations for inputs of (0,6,8 and 9). 

We can improve the method further by narrowing down the inputs we give. Rather than giving a random noise input for the corrupted case, we can give a corrupted difference in a key manner. For example, a clean input would be the image of the digit 8 and a corrupted input in this case would be the image of the digit 3. This pairing is done as the major difference between these digits is the completion of circle(s). This also aligns with the pair 5 and 6. By only working with these pairs, we create a dataset that is truly specific to the sub-task of circle detection. If an activation from a clean input (digit 8) patched onto the corrupted input (digit 3) reduces the logit difference, it means that this activation is important in detecting circles.

\subsubsection{Setup}

The setup of the experiment is as follows: 

\begin{enumerate}
\item Train the models listed in Section 2.1.4 with the original dataset. We will follow the training regime and default to the hyperparameters used by Liu et al. [1] in their code. At the end, we will have five models. 
\item We will create two subsets of the original MNIST dataset, pair one - with digits (3, 8) as (corrupted input, clean input) and pair two - with digits (5, 6). 
\item Run recursive activation patching, by Conmy et al. [2], on each of the five trained models. As described in section 2.4, our behavior (a sub-task demonstrated by the model trained to perform a main task) is the detection of circles, and our dataset for this behavior is defined in step 2. 
\item For each experiment we will output the top "k" neurons with the lowest average logit difference. These neurons form a circuit. 
\end{enumerate}

The aim of picking this task of circle detection is that a model that accurately classifies input digit images into digits, should have a circuit that is focused on detecting circles. Similarly, we can assume a circuit for detecting straight lines (for digits 1, 4, 7). 
The above set of experiments can be repeated for this task by simply forming pairs of (clean and corrupted digits like - (1, 3), (4, 9), (7, 9)). This pairing aims to have an input that makes the model exhibit the behavior (input of 4 would activate the straight-line detection circuit) and an input that is similar but will not make the model exhibit the behavior (input of 9 is close to 4 in the structural sense, but the hypothetical straight line detection circuit will not activate). This helps us perform recursive activation patching just like Conmy et al. [2]. Table 1 shows the two behaviors for our evaluation of research questions.

\begin{table}[h]
\centering
\caption{Circuit Discovery Tasks: The tasks listed are the aims of automated circuit discovery. For the first one, we are looking for a circuit in the MNIST MLP model that focuses on the detection of complete circles. The second sub-task is finding the circuit responsible for straight-line detection for the same model. For the activation patching, we need a clean input that produces the expected behavior and a corrupted input that is different in a key way which leads it to not producing the behavior.}
\label{my-label}
\begin{tabular}{@{}lllll@{}}
\toprule
\textbf{Sub Task of Circuit}             & \textbf{Input Type}   & \textbf{Example Input}   & \textbf{Expected Behaviour}  & \textbf{Metric}  \\ \midrule
\multirow{2}{*}{Circle Detection}   & Clean            & Image of digit 8        &  High activation   &  Logit Difference \\
                                    & Corrupted         & Image of digit 3    & Low activation         & Logit Difference           \\  \midrule
\multirow{2}{*}{Straight Line Detection} & Clean           & Image of digit 4        & High activation        & Logit Difference                         \\
                                    & Corrupted          & Image of digit 9      & Low activation     & Logit Difference      \\ \bottomrule
\end{tabular}
\end{table}

\subsection{Bootstrapping}

Bootstrapping was used to account for variability and randomness in results. Each metric evaluation was run 50 times with a sample size (from the paired datasets) of 500 data points. Metrics were then averaged over the samples and 95\% confidence intervals were calculated to show the variance in evaluation.

\section{Results and Analysis}

This section uses the experimental procedure from the previous section to produce data and perform analysis on it. We tackle each research question separately, using summary tables and graphs to provide evidence for our analysis.

\begin{table}[h]
\centering
\caption{Circuit Discovery Metrics: Average Logit Difference (between original network and circuit discovered), Circuit Discovery Time (in seconds), and, circuit sparsity (from 0 to 1) for each model in comparison and for each sub-task (for the MNIST Digit Classification model). Bootstrapping is performed with N = 50, each bootstrap containing 500 data points, to generate the mean and confidence intervals of metrics. Bold values show the lowest Logit Difference, indicating the circuit that best captures the sub-task and discovery time, the model that takes the least time to discover the circuit. The highest sparsity values are also shown in bold as sparser models are easier to interpret and experiment with. Across, both tasks and metrics, BIMT outperforms other training methodologies. }
\label{my-label}
\begin{tabular}{@{}lllll@{}}
\toprule
\textbf{Sub Task}                      & \textbf{Model}    & \textbf{Logit Difference} & \textbf{Discovery Time} & \textbf{Circuit Sparsity} \\ \midrule
\multirow{5}{*}{Circle Detection}   & \textbf{BIMT }             & \textbf{1.2737} $\pm$ 0.1833          & \textbf{93.7614} $ \pm 0.2155 $             & \textbf{0.9859} $ \pm 0.0009  $        \\
                                    & L1+Local          & $1.7953 \pm 0.2892$    & $94.3168 \pm 0.3293$              & $ 0.9824 \pm 0.0013$                          \\
                                    & L1 Only           & $2.3412 \pm 0.4927$    & $94.3351 \pm 0.1110$                             & $0.9551 \pm 0.0003   $                      \\
                                    & L1+Swap           & $1.9539 \pm 0.2829 $      & $94.8472 \pm 0.1105 $            & 0$.9630 \pm 0.0010$                          \\
                                    & FullyDense        & $8.9437 \pm 1.9791$           & $95.3958 \pm 0.1494 $                 & $0.8635 \pm 0.0001$                          \\ \midrule
\multirow{5}{*}{Straight Line Detection} & \textbf{BIMT}          & \textbf{1.3281} $\pm 0.3006$         & \textbf{84.5346} $\pm 0.0792$           & \textbf{0.9839} $\pm 0.0013 $                         \\
                                    & L1+Local          & $1.4663 \pm 0.3158$             & $86.4752 \pm 1.0094$         & $0.9814 \pm 0.0008$       \\
                                    & L1 Only           & $1.3828 \pm 0.2701$     & $86.0731 \pm 0.2072 $         & $0.9558 \pm 0.0003 $           \\
                                    & L1+Swap           & $1.6913 \pm 0.3807$      & $86.8943 \pm 0.3221 $          & $0.9584 \pm 0.0003$         \\
                                    & FullyDense        & $7.1213 \pm 1.5863$      & $86.5299 \pm 0.1028$        & $0.8635 \pm 1.7738$   \\ \bottomrule
\end{tabular}
\end{table}

\subsection{Research Question 1: How do networks trained to be modular affect automatic circuit discovery}

Here, we document the results of circuit discovery following the above procedure for all five models under comparison. The results of automated circuit discovery are shown in Table 2. It shows how the BIMT regime can achieve the lowest levels of logit differences to other models while being faster at discovery and forming sparser circuits. 

The results are also visualized in Figure 3 and Figure 4. With the help of these results, we can falsify RQ1 hypothesis 3, which suggested that BIMT would hamper the ability to find circuits. With much higher levels of sparsity, the search space for automated circuit discovery (which iterates over every activation) reduces significantly. Thus, the faster speed of discovery supports our assertion for RQ1 hypothesis 2. We can also falsify RQ 1 Hypothesis 1, Modularity can create sparser models with lower logit differences. These results are consistent across the bootstrap samples and the two behaviors we examined. 

\subsubsection{Quality of Circuits}

Logit Difference is the difference between the logit output of the original model and the logit output of the circuit formed. The goal of using this is to replicate the output of the original model for the specific sub-task at hand. Let's quickly do a few inference tests to contextualize this. 

The circle detection circuit formed by BIMT, when inferenced on the task-specific dataset (of digits 3 and 8) outputs an accuracy of 97.04\% while the accuracy of the circuit formed by the L1-only model is 96.08\%. Figure 2 shows the discovered circuit for BIMT and L1-only models. It showcases their average logit score with the percentage sparsity they were able to achieve.

While these are comparable, the goal of the research is not to create the best possible circuit but rather to evaluate the different effects modular training can have on automated circuit discovery. The method of circuit discovery can be improved and fine-tuned for this task to get better models but the concept of recursive activation patching is what is important to us.

\subsubsection{Discovery Time}

The time taken to discover a circuit might be one of the most important metrics in this research. It aims to find a training regime that can boost and potentially upscale current mechanistic interpretability techniques to reach large language models. 

From Table 2 and Figure 4, we see that BIMT can discover circuits with the smallest times. A model trained to be modular is much sparser than the rest, so the search space for recursive iterations lessens significantly. This is one of the biggest factors contributing to its speed of discovery. One concern was the trade-off it would have with the quality of the circuits, but from our results, that is not the case.

\begin{figure}
    \centering
    \includegraphics[scale=0.6]{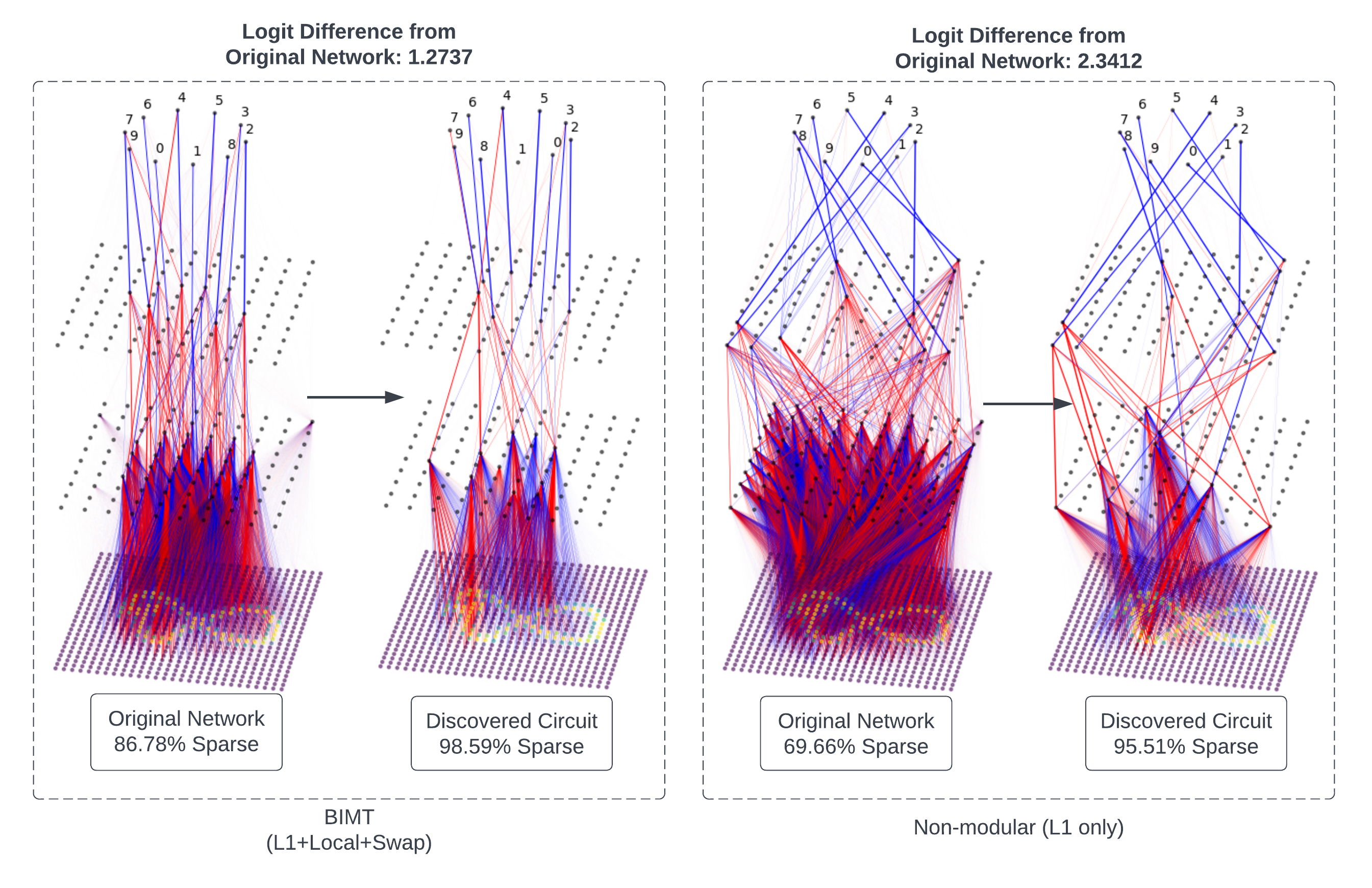}
    \caption{Comparison between original network and discovered circuits for BIMT and L1 Only model. We notice that on inference, BIMT can produce significantly lower logit differences. Red edges denote positive edges, while blue edges denote negative ones. Discovered circuits also display modularity in the case of BIMT. }
    \label{fig:logits}
\end{figure}

\begin{figure}
    \centering
    \includegraphics[scale=0.6]{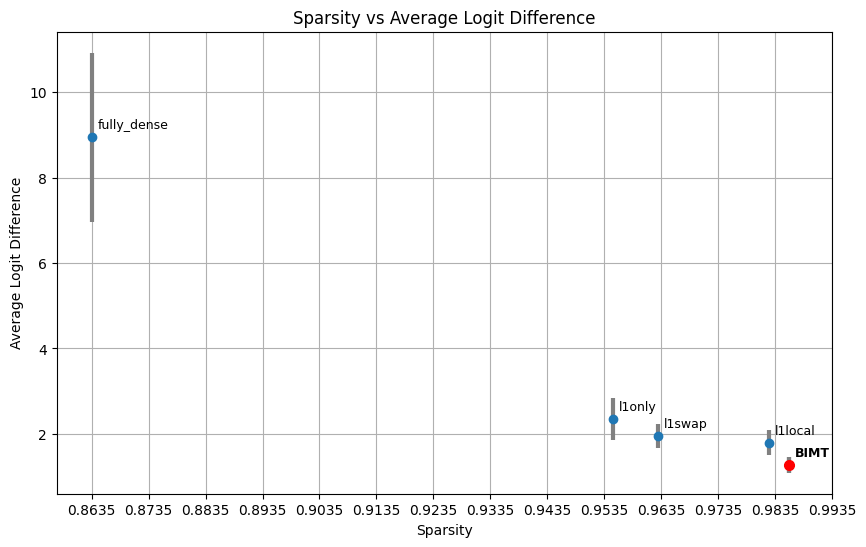}
    \caption{Comparison of the sparsity of discovered circuit vs average logit difference. Each data point represents a model under comparison. The gray vertical lines represent the 95\% confidence interval calculated by bootstrapping. Models in the bottom right are highly valuable, as they represent a low average logit score (showing the presence of behavior) and a high sparsity of models (avoiding redundancy in circuits). BIMT can consistently provide the highest sparsity and lowest logit differences for the task of circle detection.}
    \label{fig:logits}
\end{figure}

\begin{figure}
    \centering
    \includegraphics[scale=0.6]{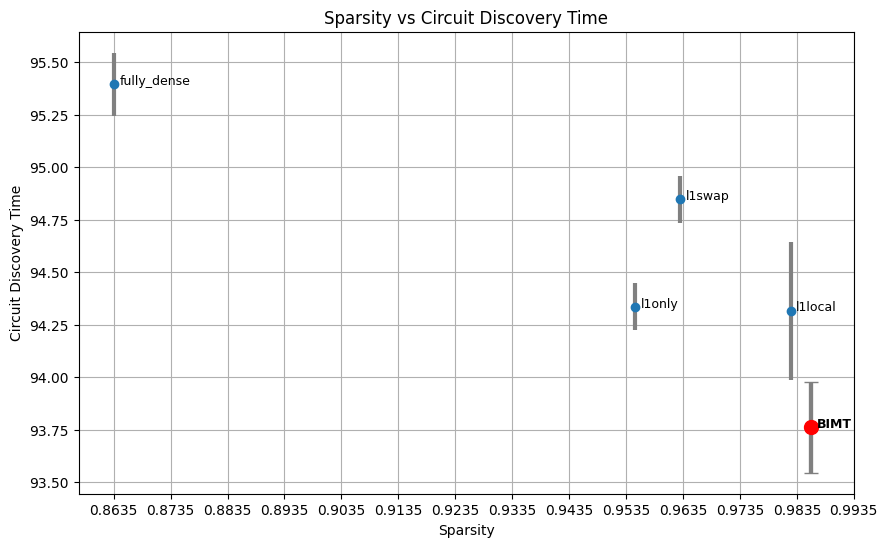}
    \caption{Comparison of the sparsity of discovered circuit vs time taken for discovery. Each data point represents a model under comparison. The gray vertical lines represent the 95\% confidence interval calculated by bootstrapping. BIMT can consistently discover sparser circuits with the lowest mean time of discovery. As BIMT is the sparser network to begin with, the discovery time decreases.}
    \label{fig:logits}
\end{figure}

\subsubsection{Sparsity of Discovered Circuit}

The sparsity of the model is a big concern for us. Ideally, a sparse model is easier to experiment with. Figure 2 demonstrates the sparsity of the network and the discovered circuit for BIMT and L1 Only models. We can see from this and the x-axis of Figures 3 and 4, that networks that are already sparse, create circuits that are sparser. 

\subsection{Research Question 2: How does the computational efficiency vary with respect to modularity?}

To evaluate the second research question, bookkeeping of various metrics was carried out throughout the experimental procedure of the first research question.  

While training each model, we made sure to log the time required to complete the training regime and GPU memory used. The latter was calculated with a Python library called "subprocess". Inference time for each model was also bootstrapped to account for variability with the same number of samples as the other experiments. 

\subsubsection{Training Memory Allocation}

The results of memory allocation during training are shown in Figure 6. Modular training resulted in a much higher memory requirement than other models. We hypothesized that storing coordinate values would lead to a rise in memory allocation. However, through our experiments, the operation of "swap" - (swapping neurons that reduce the distance-dependent loss) causes the most significant rise in memory allocation. 

\begin{figure}
    \centering
    \includegraphics[scale=0.5]{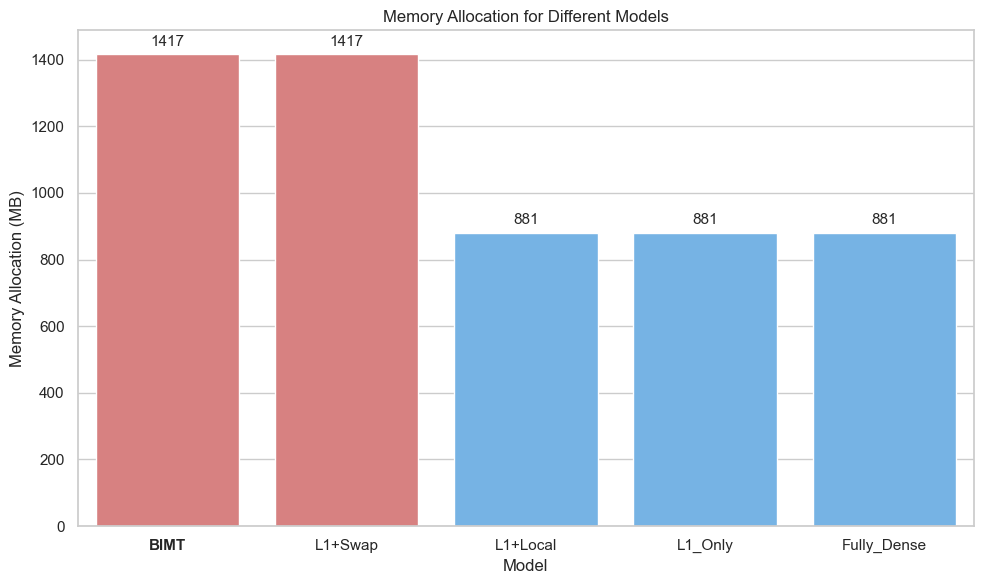}
    \caption{ Memory allocated to CUDA GPU during training. Models with "swap" require almost 1.5 times the memory to train in comparison to other models.}
    \label{fig:logits}
\end{figure}

\subsubsection{Training and Inference Times}

The swap operation also significantly increases the training times for a model as shown in Figure 7. This highly supports our RQ2 hypothesis 2, however, the reason for the increase in training time is the "swap" operation and not the additional "distance calculation". We perform the same bootstrapping methodology from section 3.3 for evaluating inference times for each model. From Figure 8, we can see that the lowest inference times are achieved by L1 only model. Adding functionalities like swapping, and distance-dependent regularization increases the inference time. With this result, we can falsify hypothesis 3, the model training regime does affect inference times. 

To put the results of Figure 8 in context, inferencing 500000 images for the BIMT model will take an average of 65.77 seconds, while inferencing the same for L1 only model will take an average of 64.65 seconds. Though the difference might not seem much, the models under consideration are far simpler than LLMs.

\begin{figure}
    \centering
    \includegraphics[scale=0.5]{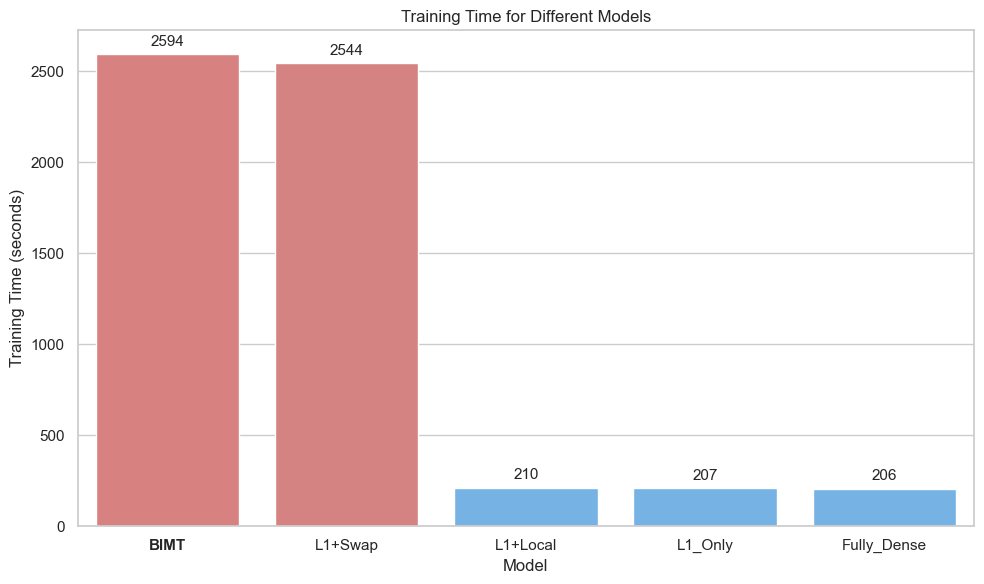}
    \caption{Training time (in seconds - for the entire training procedure) for each model under comparison. We notice that swaps drastically increase the training time for models. The time in seconds is shown above the bar for each model.}
    \label{fig:logits}
\end{figure}

\begin{figure}
    \centering
    \includegraphics[scale=0.6]{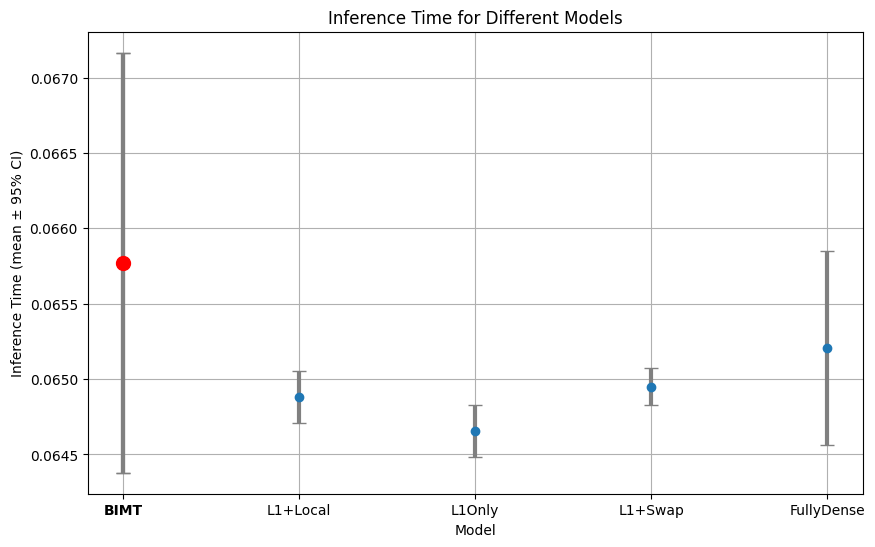}
    \caption{Inference time (in seconds) for each model under comparison, with gray vertical lines representing the 95\% confidence interval calculated by bootstrapping. The lowest mean inference time is achieved by the L1-only model. With each functionality addition (like swap, local regularization}, increasing the inference time. 
    \label{fig:logits}
\end{figure}

\section{Related Work} 

\textbf{Mechanistic Interpretability} (MI) in neural networks, inspired by cellular biology, aims to demystify complex models by analyzing their sub-networks or circuits. This concept, initially explored in image networks by Olah et al. [4], emphasizes understanding network functionalities through these circuits. Works on interpretability by Ribeiro et al. [8], Selvaraju et al. [9], and Kim et al. [10] have been mainly focused on the explanation of the model decisions. Instead, Olah [7] introduces MI with the aim of understanding the internal information processing of a model. Elhage et al. [11] formulate a mathematical framework for understanding transformer architectures and Nanda et al. [12], Cammarata et al. [13], and Wang et al. [14] explore and present circuits in various models. Bau et al. [16] and Meng et al. [3] have focused on subgraphs of models to edit a model’s sub-task and test interpretability hypotheses. 

\textbf{Scaling and automation} Before the research of Conmy et al. [2], automation of interpretability was not explored besides Bills et al. [15] who used language models to label neurons in language models. Wu et al. [17] use causal testing to propose a method to automatically test if neural networks are implementing certain algorithms. The work by Conmy et al. [2] was focused on discovering the algorithm implemented by a model. They propose the MI technique of recursive activation patching for automatic circuit discovery, a technique heavily utilized in this paper. 

\textbf{Modularity} Research by Pfeiffer et al. [18] suggests that modularity in models can help generalization in transfer learning. The concept of modularity, while not new, hasn't been sufficiently explored concerning modern MI techniques. The paper by Filan et al. [19] found that trained networks exhibit more cluster-ability than randomly initialized ones. The work by Liu et al. [1] formally introduces a biologically inspired training regime for forming modules in networks during training. More recent work by Liu et al. [20] showcases how this training regime can help RNNs exhibit brain-like anatomical modularity.

To summarize, we saw that most of the focus has been shifted to Transformer and GPT-style networks rather than a general model. Most recent research focuses almost exclusively on LLMs with the hopes of solving alignment. However, even with all this focus, MI seems to be very limited for large-scale networks (as most experiments use GPT-2-Small). This is largely due to the lack of automation which is starting to change slowly as the frontier moves forward. The work by Conmy et al. [2] is the first step in fixing that limitation. Not much is known about how modularity would affect automated interpretation. 

\section{Conclusion}

In this study, we have thoroughly examined the Brain Inspired Modular Training (BIMT) method and its impact on the interpretability of neural networks. BIMT is inspired by the organizational principles of biological brains and seeks to enhance the mechanistic interpretability of artificial neural networks by introducing modularity through a geometrically constrained training process. Our findings present a strong case for BIMT as a method that significantly improves the efficiency of Automated Circuit Discovery.

Our research directly addresses two fundamental questions. Firstly, we investigated how modularity influences the discovery of circuits within neural networks. We demonstrated that BIMT allows for the efficient identification of sparse, interpretable circuits, thereby reducing the complexity of the discovery process. This is evidenced by the low logit differences, faster discovery times, and higher sparsity levels achieved by BIMT compared to other models under comparison. These results suggest that BIMT enables more straightforward and rapid interpretability, which is critical for the analysis of large-scale models such as GPT4 or LLAMA.

Secondly, we explored the computational efficiency of BIMT in terms of memory allocation and inference speed. While BIMT does require more memory during training, largely due to the neuron-swapping operation, this is offset by the benefits it provides in interpretability and discovery speed. The slight increase in inference time is a reasonable trade-off, given the interpretability advantages that BIMT brings to the table. 

In conclusion, BIMT emerges as an innovative and practical approach to neural network interpretability, aligning with the urgent need for transparent and reliable AI systems. Our pioneering work in comparing BIMT for Automated Circuit Discovery contributes to the critical discourse on AI interpretability and sets the stage for future research to refine and apply this methodology to more complex AI systems.

\subsection{Threats to Validity}

The results demonstrated in this report suffer from various internal and external threats to validity. To account for results that are specific to circle detection, another task of straight-line detection was added, leading to similar results. However, all of the testing was still only done on the MNIST dataset and MLP model architecture. Adding multiple tasks and bootstrapping helped us in evaluating the generalizability of results but the threats remain.

Additionally, we would want our sub-network explaining a feature of the actual model to be as sparse as possible. The Automated Circuit Discovery paper by Conmy et al. [2] assumed that the smaller the circuit, the easier it will be to interpret it. However, recent research in Mechanistic Interpretability by Elhage et al. [4] showcases the existence of "Superposition". This is a phenomenon when models represent more features than they have dimensions. When features are sparse, superposition allows compression beyond what a linear model would do, at the cost of "interference" that requires nonlinear filtering. This is a big problem as the entire aim of the Principal Paper by Liu et al. [1] is to prune and regularize the architecture while training to create modules. The biggest threat to interpretability comes from the fact that training the network to be more modular might sparsify it to an extent where the neurons are all "polysemantic" - neurons that respond to unrelated mixtures of inputs. If this is true, then discovering sub-networks will not directly lead to interpretable modules.

\section{Future Work}

The work done in the research allowed us to evaluate BIMT with other training regimes in a fair context and present additional computational metrics for it. The work aimed to show the working of the system in a system that is well established and provide evidence that it could extrapolate to larger systems. Future work can build on these results in various ways:

\begin{itemize}
    \item The BIMT architecture can be tested on transformer networks, which was one of the future works of the paper by Liu et al. [1]. 
    \item Experiments can be done on superposition [4] and the effect of modularity on it. 
    \item Recent work by Syed et al [6], suggests that attribution patching is an effective and faster approximation to activation patching and that it outperforms recursive activation patching. Evaluation of this method can be used instead of activation patching for larger networks. 
    \item This work specifically focuses on the image (digit) classification task. Evaluation can be done on other tasks for the original networks. 
\end{itemize}

\section*{Acknowledgments}

We would like to express our sincere gratitude to Prof. David Jenson for his invaluable guidance and profound insights that greatly shaped this research. We are also thankful to two computer science students from the University of Massachusetts Amherst who, although remaining anonymous, provided critical feedback that significantly improved the quality of this work. 

\section*{References}
\medskip
{
\small

[1] Liu, Ziming, Eric Gan, and Max Tegmark. "Seeing is Believing: Brain-Inspired Modular Training for Mechanistic Interpretability." arXiv preprint arXiv:2305.08746 (2023).

[2] Conmy, Arthur, et al. "Towards automated circuit discovery for mechanistic interpretability." arXiv preprint arXiv:2304.14997 (2023).

[3] Meng, Kevin, et al. "Locating and editing factual associations in GPT." Advances in Neural Information Processing Systems 35 (2022): 17359-17372.

[4] Elhage, et al., "Toy Models of Superposition", Transformer Circuits Thread, 2022.

[5] Nanda, Neel. "Attribution patching: Activation patching at industrial scale." URL: https://www. neelnanda. io/mechanistic-interpretability/attribution-patching (2023).

[6] Syed, Aaquib, Can Rager, and Arthur Conmy. "Attribution Patching Outperforms Automated Circuit Discovery." arXiv preprint arXiv:2310.10348 (2023).

[7] Chris Olah. Mechanistic interpretability, variables, and the importance of interpretable bases, 2022. 

[8] Ribeiro, Marco Tulio, Sameer Singh, and Carlos Guestrin. "" Why should i trust you?" Explaining the predictions of any classifier." Proceedings of the 22nd ACM SIGKDD international conference on knowledge discovery and data mining. 2016.

[9] Selvaraju, Ramprasaath R., et al. "Grad-cam: Visual explanations from deep networks via gradient-based localization." Proceedings of the IEEE international conference on computer vision. 2017.

[10] Kim, Sunnie SY, et al. "HIVE: Evaluating the human interpretability of visual explanations." European Conference on Computer Vision. Cham: Springer Nature Switzerland, 2022.

[11] Elhage, Nelson, Neel Nanda, Catherine Olsson, Tom Henighan, Nicholas Joseph, Ben Mann, Amanda Askell, Yuntao Bai, Anna Chen, Tom Conerly, Nova DasSarma, Dawn Drain, Deep Ganguli, Zac Hatfield-Dodds, Danny Hernandez, Andy Jones, Jackson Kernion, Liane Lovitt, Kamal Ndousse, Dario Amodei, Tom Brown, Jack Clark, Jared Kaplan, Sam McCandlish, and Chris Olah (2021). “A Mathematical Framework for Transformer Circuits”. In: Transformer Circuits Thread. URL: https://transformer-circuits.pub/2021/framework/index.html.

[12] Nanda, Neel, Lawrence Chan, Tom Lieberum, Jess Smith, and Jacob Steinhardt (2023). “Progress measures for grokking via mechanistic interpretability”. In: The Eleventh International Conference on Learning Representations. URL: https://openreview.net/forum?id=9XFSbDPmdW.

[13] Cammarata, Nick, Gabriel Goh, Shan Carter, Chelsea Voss, Ludwig Schubert, and Chris Olah (2021). “Curve Circuits”. In: Distill. https://distill.pub/2020/circuits/curve-circuits. DOI: 10.23915/
distill.00024.006.

[14] Wang, Kevin, et al. "Interpretability in the wild: a circuit for indirect object identification in gpt-2 small." arXiv preprint arXiv:2211.00593 (2022).

[15] Bills, Steven, et al. "Language models can explain neurons in language models." URL https://openaipublic. blob. core. windows. net/neuron-explainer/paper/index. html.(Date accessed: 14.05. 2023) (2023).

[16] Bau, David, Steven Liu, Tongzhou Wang, Jun-Yan Zhu, and Antonio Torralba (2020). Rewriting a Deep Generative Model. URL: https://arxiv.org/abs/2007.15646.

[17] Wu, Zhengxuan, et al. "Interpretability at scale: Identifying causal mechanisms in alpaca." arXiv preprint arXiv:2305.08809 (2023).

[18] Pfeiffer, Jonas, et al. "Modular deep learning." arXiv preprint arXiv:2302.11529 (2023).

[19] Filan, Daniel, et al. "Clusterability in neural networks." arXiv preprint arXiv:2103.03386 (2021).

[20] Liu, Ziming, et al. "Growing Brains: Co-emergence of Anatomical and Functional Modularity in Recurrent Neural Networks." arXiv preprint arXiv:2310.07711 (2023).

}

%%%%%%%%%%%%%%%%%%%%%%%%%%%%%%%%%%%%%%%%%%%%%%%%%%%%%%%%%%%%

\end{document}